# Machine Learning for Robust Identification of Complex Nonlinear Dynamical Systems: Applications to Earth Systems Modeling

Nishant Yadav
Sustainability & Data Sciences Lab
Civil & Environmental Engineering
Northeastern University
Boston, MA, USA
yadav.ni@northeastern.edu

Sai Ravela
Earth Signals and Systems Group
Earth, Atmospheric and Planetary
Sciences, Massachusetts Institute of
Technology (MIT)
Cambridge, MA, USA
ravela@mit.edu

Auroop R. Ganguly†
Sustainability & Data Sciences Lab
Civil & Environmental Engineering
Northeastern University
Boston, MA, USA
a.ganguly@northeastern.edu

## ABSTRACT

Systems exhibiting nonlinear dynamics, including but not limited to chaos, are ubiquitous across Earth Sciences such as Meteorology, Hydrology, Climate and Ecology, Engineering such as chemical reactions and structural dynamics, as well as Biology such as neural and cardiac processes. However, System Identification remains a challenge. Thus, in climate and earth systems models, while governing equations follow from first principles and understanding of key processes has steadily improved, the largest uncertainties are often caused by parameterizations such as cloud physics, which in turn have witnessed limited improvements over the last several decades. Climate scientists have pointed to Machine Learning enhanced parameter estimation as a possible solution, with proof-of-concept methodological adaptations being examined on idealized systems. While climate science has been highlighted as a "Big Data" challenge owing to the volume and complexity of archived model-simulations and observations from remote and in-situ sensors, the parameter estimation process is often relatively a "small data" problem. The latter is caused by multiple interacting factors such as inadequate data and imperfect physics at high-enough resolutions, limited historical records before the dawn of remote sensors such as earth-observing satellites and weather radars. A crucial question for data scientists in this context is the relevance of state-of-the-art data-driven approaches including those based on deep neural networks or kernel-based processes. Here we consider a chaotic system – two-level Lorenz-96 – used as a benchmark model in the climate science literature [6], adopt a methodology based on Gaussian Processes for parameter estimation and compare the gains in predictive understanding with a suite of Deep Learning and strawman Linear Regression methods. Our results show that adaptations of kernel-based Gaussian Processes can outperform other approaches under small data constraints along with uncertainty quantification; and needs to be considered as a viable approach in climate science and earth system modeling.

## KEYWORDS

Nonlinear Dynamics, Climate Modeling, Gaussian Processes, Deep Learning

## 1 Introduction

With the advent of big data, machine learning and data science has ushered in a new era of predictive understanding of complex, high-dimensional data in problems like image classification and speech recognition [1, 2]. Given massive datasets, state-of-the-art deep learning architectures can be trained to act as universal functional approximators to achieve results which sometimes even exceed human accuracy. For example, ResNet, which won the ImageNet challenge in 2015, was trained on 1.2 million images [3]. However, many fundamental problems in science deal with "small data" where data availability is limited, and simulated data is difficult to generate due to high computational cost or may even be infeasible due to incomplete understanding of the underlying process physics. While a simple online search returns thousands of pictures of a particular object, and millions of Wikipedia articles are downloaded in seconds, collecting a single run of a high-resolution climate model is both time-consuming and expensive. Such a critical dependency on large datasets has limited the success of machine learning in problem spaces where data is hard to come by.

Nonlinear dynamical (NLD) systems are ubiquitous in nature with wide applications ranging from fluid dynamics, biomedical signal processing (e.g., ECG), epidemiology and climate modeling [4-6] among others. However, the sheer complexity of the system may render a first-principles modeling approach infeasible. Instead, data-driven methods provide an alternative to discover the governing equations from observations. These systems are mathematically defined using a set of coupled differential equations. In its simplest form, a dynamical system is of the form:

$$\dot{x} = f(x, u)$$
$$y = h(x) + \eta$$

The vectors $x$ and $u$ denotes the state of the system and input, respectively, at time t and the function $f(.)$ defines the dynamic constraints that define the system including parametrization. The vector $y$ refers to the measured observations and $h$ is the transformation mapping states to observations, and $\eta$ is the measurement noise. Depending on whether the model structure is known (black-box vs white box modeling), the goal of system identification is to estimate either the model itself or the model parameters. In this work, we assume that the model structure is known as is the case in climate modeling.

The focus in this work is on NLD systems and parameter estimation in the context of earth system/climate modeling. To this end, we use a benchmark chaotic system – two-level Lorenz-96 (L96-2L) – developed by Lorenz [7] representative of the general circulation of the atmospheric and exhibiting similar properties such as a chaotic error growth rate and multiscale interaction. It consists of a coupling of variables evolving over slow and fast timescales (discussed in detail in Section 3.1). It has served as a testbed for machine learning research in parameter estimation for more complex actual climate models [8, 9, 38].

The specific goal of this paper is twofold: first, to derive probabilistic estimates of the model parameters of the L96-2L system using Gaussian Process Regression (GPR); second, sampling from the estimated parameter distribution, the system attractor is then replicated and compared with the true attractor. A suite of shallow and deep learning algorithms are used as benchmarks to compare the relative performance under the small data constraint as discussed above.

The remainder of the paper is organized as follows. Section 2 discusses some of the recent works in NLD system identification. Section 3 presents the background (methodology and data) necessary for this study. In Section 4 (Results) we compare our method with two deep learning methods and linear regression as baselines. Section 5 outlines a discussion on the results, limitations and future work.

## 1.1 Relevance to Climate Modeling

Earth System Models (ESMs), previously known as Global Climate Models (GCMs), have become useful tools for climate science as well as for stakeholder and policy communities (e.g., Intergovernmental Panel for Climate Change [23]; United Nations Sustainable Development Goals. However, despite advances in our understanding of the relevant physics and biogeochemistry, along with developments in computational power and availability of Big Data from remote sensors and archived model simulations, crucial knowledge-gaps remain. While the fundamental structures of ESMs are often based on first-principles physics encapsulated within partial differential equations, key processes like cloud physics rely on simplified parameterizations [39, 40] owing to uncertainties in physics [53], inadequacy of computational resolutions, and limitations of data availability. The challenges in parameterizations can cause major uncertainties [50] in predictive understanding, as exhibited by the latest generation ESMs rather prominently (e.g., in the scientific literature [24, 41] and even in the media [25]).

Climate and earth system modelers have recently pointed to the potential value of Machine Learning, including Deep Learning in parameter estimation or crucial processes such as cloud convention where long-standing knowledge gaps continue to exist. A paper in the journal Geophysical Research Letters [6] proposed "Earth System Modeling 2.0" and showcased (with the 96-2L model as a proxy for ESMs) how Machine Learning can help in parameter estimation. A perspective article in the journal Nature [21] and an editorial article in Science magazine [22] pointed to the challenges and the opportunities. A recent paper [8] in the journal Proceedings of the National Academy of Sciences presented a Deep Learning approach to represent sub-grid cloud processes. A schematic of how Machine Learning can inform ESMs – along with the connection to an idealized Nonlinear Dynamical System (specifically, the L96-2L) often used as a proxy model – is shown schematically in Figure 1.

## 2 Related Work

We approach System Identification as parameter estimation, which is as such a well-developed subject []. Classical estimation approaches include variational solutions to two-point boundary value problems (2BVP) [10], the recursive Kalman and extended Kalman filters, and the Rauch-Tung-Striebel smoother [11]. Practical Bayesian joint state-parameter estimation with uncertainty quantification for nonlinear high-dimensional systems have emerged in the form of efficient ensemble filters [27], fast ensemble smoothers [28] and Particle Filters [29], with numerous variants. Probabilistic Graphical Models unify Bayesian estimation for both random fields and stochastic processes and many Bayesian inference algorithms can be reduced to a form of Belief Propagation [30].

More recently, in the ML/DL ("Learning") space, researchers have used reservoir computing [12, 13], LSTMs [14], Random Forests [4], multi-step Deep Neural Networks [15, 49] for forecasting the future states in different canonical problems in NLD using only the past observations. Although the above methods show good predictive performance, the methods do not capture the true governing mechanism of the system; in the sense that model parameters are not estimated.

Although Learning is itself often a parameter estimation problem, our interest is in its use for parameter estimation of dynamical systems, in particular through the use of ensemble and deep learning, and kernel machines. Although some approaches for dynamical systems unrelated to present work including Ensemble Learning [31] and Information-theoretic Learning alternatives [32] have been proposed, learning in parameter estimation is not well developed. Gaussian Processes (GPs) [35] as a learning method proposed here are closely related to Gaussian Graphical models (the update equations are similar, but the formulations are different) [33]. GPs are also related to deep learning [26] in the sense of an infinite-width limit.

In this work, instead of directly predicting future states, we first estimate the model parameters which combined with the model structure provides us the complete system information to simulate new data for different initial conditions. [16]. We draw motivation from the recent works of Raissi et al. [17] that have used Gaussian Processes (GPs) for solving partial differential equations. GPs offer several practical advantages: 1) Intrinsic Bayesian approach captures model uncertainty [52] by providing confidence bounds over the estimated parameters 2) ability to incorporate prior domain knowledge by selecting appropriate kernel functions  3) Occam's

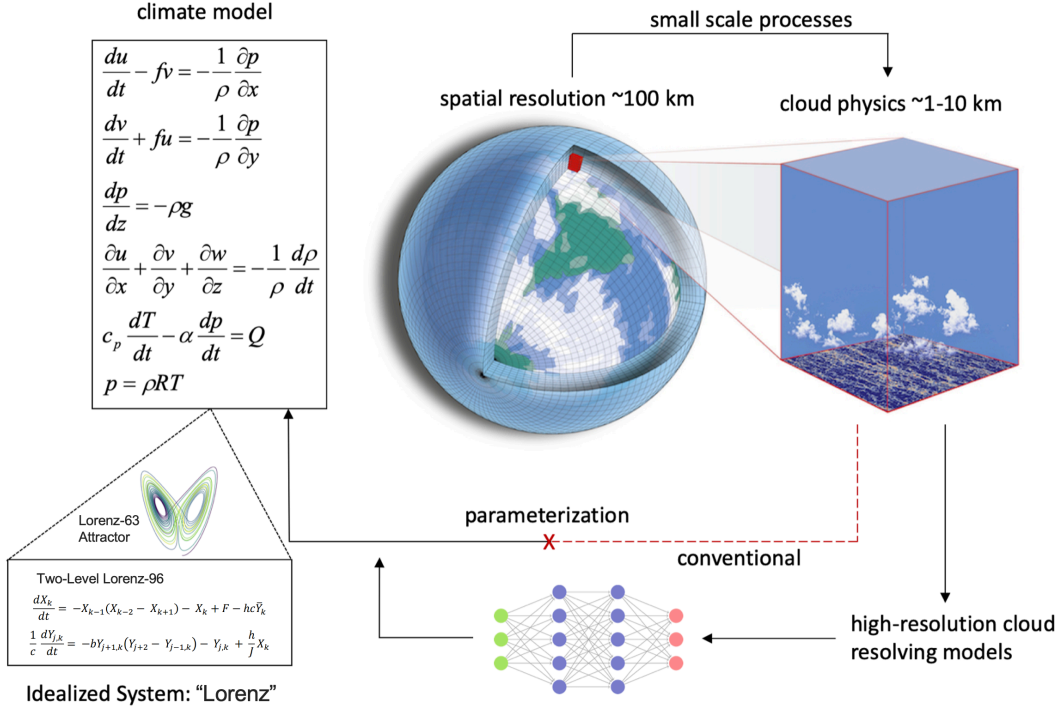

**Figure 1: Schematic diagram showing the parameterization problem in coarse global climate models. Machine Learning based parameterization method is proposed to replace traditional heuristic parameterization schemes. Inset shows the two-level Lorenz-96 system (L96-2L) used as a simplified testbed model as suggested in emerging literature (ESM 2.0 [6]). The butterfly-shaped Lorenz-63 attractor is shown as a representational image for a NLD (chaotic) system.**

Razor [47] automatically incorporated by the marginal likelihood function - a more complex model can account for many more data sets than a simple model, but the probabilities have to integrate to unity, thus, more complex models are automatically penalized more. On the other hand, GPs are caveated by its cubic computational complexity. The inversion of the $n \times n$ covariance matrix requires a memory complexity of $O(n^2)$ and computational complexity of $O(n^3)$. This cost becomes prohibitive for large training data. Decreasing the computational complexity of GPs is an active research area with different algorithms available that extract low-rank approximation [46, 51] for the covariance matrix. For example, a rank-m Cholesky factorization can be computed in $O(nm^2)$ time. Since the central idea of this work is to demonstrate GPs on small data, we have not looked into reducing the time complexity.

## 3 Background

### 3.1 Two-Level Lorenz-96 System

The two-level Lorenz-96 [7] is a prototype model of the mid latitude atmosphere. The model describes the time evolution of the components $X_j$ of a spatially discretized atmospheric variable over a single latitude circle. Associated with each $X_j$ are $Y_k$ variables representing unresolved subgrid processes (e.g. cloud microphysics).

$$\frac{dX_k}{dt} = -X_{k-1}(X_{k-2} - X_{k+1}) - X_k + F - hc\bar{Y}_k \quad [1]$$

$$\frac{1}{c}\frac{dY_{j,k}}{dt} = -bY_{j+1,k}(Y_{j+2} - Y_{j-1,k}) - Y_{j,k} + \frac{h}{J}X_k \quad [2]$$

where,

$$\bar{Y}_k = \frac{1}{J}\sum_{j=1}^{J} Y_{j,k}$$

This set of equations are coupled through the mean term $\bar{Y}_k$ and this coupling is controlled by three keys parameters: $b, c$ and $h$. The parameter $b$ controls the amplitude of the nonlinear interactions among the fast variables, while the parameter $c$ controls how rapidly the fast variables fluctuate relative to the slow variables and the parameter $h$ controls how strong the coupling between the fast and slow variables is. Lorenz-96 is commonly used as a benchmark model for weather and climate prediction as well as recently in ML based parametrization schemes for Earth System Models.

## 3.2 Gaussian Processes

**Definition 1:** *A Gaussian process is a collection of random variables, any finite number of which have a joint Gaussian distribution.* [18]

A Gaussian process (GP) is completely specified by its mean function $m(x)$ and covariance function $k(x, x')$. This is a natural generalization of the Gaussian distribution whose mean and covariance is a vector and matrix, respectively. The Gaussian distribution is defined over finite-dimension vectors, whereas the Gaussian process is over functions, discrete or continuous. We can write:

$$f = GP(m, k)$$

i.e. the function $f$ is distributed as a GP with a mean function $m$ and covariance function $k$. The covariance kernel function used in this work is the squared exponential kernel [34].

Now, for any finite subset $X = \{x_1, x_2, \ldots x_n\}$ in the domain of $x$, the marginal distribution $f(X)$ is a multivariate Gaussian Distribution (definition 3.1):

$$f(X) = N(m(X), k(X, X))$$

with mean $\mu = m(X)$ and covariance matrix $\Sigma = k(X, X)$

Let $f$ be the known function values of the training cases and let $f*$ be a set of function values corresponding to the test set inputs, $X*$. we write out the joint distribution of everything we are interested in:

$$\begin{bmatrix} f \\ f^* \end{bmatrix} = N\left(\begin{bmatrix} \mu \\ \mu_* \end{bmatrix}, \begin{bmatrix} \Sigma & \Sigma* \\ \Sigma*^T & \Sigma** \end{bmatrix}\right)$$

where we've introduced the following shorthand: $\mu = m(x_i), i = 1, \ldots, n$ for the training means and analogously for the test means $\mu_*$; for the covariance we use $\Sigma$ for training set covariances, $\Sigma*$ for training-test set covariances and $\Sigma**$ for test set covariances. Since we know the values for the training set $f$ we are interested in the conditional distribution of $f*$ given $f$ which is expressed as:

$$f^*|f \sim N(\mu_* + \Sigma_*^T \Sigma^{-1}(f - \mu), \quad \Sigma_{**} - \Sigma_*^T \Sigma^{-1} \Sigma_*)$$

This is the posterior distribution for a set of test cases. The corresponding posterior process is:

$$f^* \mid D \sim GP(m_D, k_D)$$
$$m_D(x) = m(x) + \Sigma(X, x)^T \Sigma^{-1}(f - m)$$
$$k_D(x, x') = k(x, x') - \Sigma(X, x)^T \Sigma^{-1} \Sigma(X, x')$$

## 4 Problem Formulation

The regression problem here is to estimate the coupling parameter $h$ as a function of the resolved, large-scale variable X. Since all $X_i$ are statistically similar, considering only $X_1$ should suffice. Each input datapoint is a length $n$ temporal "snapshot" extracted from the $X_1$ time series (see Data Generation). The corresponding target is the value of parameter $h$ used to generate the $X_1$ trajectory.

To estimate: $f^*{}_h | f_h$, with $f_h \sim GP(m(X_1'), k(X_1', X_1'))$

where $X_1'$ is a finite subset of $X_1 = \{X_1(t) \ldots . . X_1(t + n\Delta t)\}$ available as training data.

## 5 Experiments

**Data Generation:** 8 experiments with 200 simulation each for different combinations of the parameters $b, c$ and $h$ are considered. For each simulation, we solve equations [1] and [2] using a Fourth Order Runge-Kutta numerical solver with a step $\Delta t = 0.005$ up 4100 timesteps. Accounting for transient effects, the first 1000 points are discarded resulting in a time series of 4000 point. From here on, we express $(1 / \Delta t)$ timesteps as 1 Model Time Unit (MTU) [36, 37]. In total, 1600 such time series are generated.

The eight experiments differ by varying the number of slow (X) and fast variables (Y), the forcing F and the temporal regime from where the training data is extracted. A higher forcing F results in more chaotic behavior of the L96-2L system, while the two training data regimes correspond to low and high temporal variability in the trajectories of the $X_i$. The details are described below:

**Table 1: Summary of Experiments**

| Experiment # | X | Y | F | Training Regime |
|---|---|---|---|---|
| 1 | 4 | 4 | 10 | 5 < MTU < 7 |
| 2 | 4 | 4 | 10 | 15 < MTU |
| 3 | 4 | 4 | 20 | 5 < MTU < 7 |
| 4 | 4 | 4 | 20 | 15 < MTU |
| 5 | 8 | 8 | 10 | 5 < MTU < 7 |
| 6 | 8 | 8 | 10 | 15 < MTU |
| 7 | 8 | 8 | 20 | 5 < MTU < 7 |
| 8 | 8 | 8 | 20 | 15 < MTU |

**Data Preprocessing:** To emulate a real scenario of availability of only partial observations, only the resolved (observed) variable X is considered as input for training data. The unresolved variable Y represents sub-grid processes in a climate model are often not available as observations from coarse-resolution climate models.

From each $X_1$ time series, 5 temporal snapshots of length *n*=10 are extracted as the training data. Thus, shape of the training data for each experiment is [simulations=200; no. of datapoints per simulation=5; length of each datapoint = 10] which is reshaped into a [1000, 10] NumPy array. The corresponding target shape is [1000, 1]. For different experiments (such as #1 and #2), all else being equal, the temporal "snapshots" are generated from two different regimes in the $X_i$ time series. For 5<MTU<7, the variation in $X_i$ is less drastic, while MTU > 15 is considered to capture greater chaotic behavior and compare the predictive performance of the models.

# 6 Results

Performance metrics for the 8 experiments are presented in Table 2 to Table 9. The metrics used are:

- Mean Squared Error (MSE)
- Mean Absolute Error (MAE)
- Coefficient of Determination $R^2$ – provides a measure of the proportion of total variation of outcomes explained by the model.

$$R^2 = 1 - \frac{\sum_i (\hat{y}_i - \bar{y}_i)^2}{\sum_i (y_i - \bar{y}_i)^2}$$

- Bhattacharya Distance – used to measure similarity between the inferred and true probability distribution of the estimated parameter (h). For two probability distributions, $p(x)$ and $q(x)$, it is calculated as:

$$D_B = -\ln\left(\sum_{x \in X} \sqrt{p(x)q(x)}\right)$$

- Pearson's Correlation

**Baseline Models** used to compare the performance of GP:
- FC-DNN – a 4-layer [64, 32, 16, 8, 1] fully connected deep neural network
- Stacked-LSTM – a 2-layer stacked LSTM with [64, 32] hidden units
- Linear Regression (LR)

## 6.1 Evaluation

**Performance Metrics:**
- In the less variable training regime (< 5 MTU < 7), Gaussian Process Regression (GPR) consistently outperforms across all metrics, for both levels of chaotic behavior (Table 2 to Table 5). Between the two DL models, FC-DNN performs no better than a simple linear regression; and the stacked-LSTM performs worse.
- For the MTU > 15 training regime (Table 6 to Table 9), predictive power of GPR deteriorates while it remains roughly the same for the two DL models. GPR still marginally outperforms FC-DNN for the less chaotic (F=10) Lorenz-96. However, the performance of GP, FC-DNN and stacked-LSTM converge as chaoticity increases, because the intrinsic dimensionality of the system attractor increases and the system inherently unpredictable. For F=20 and MTU > 20 training regime, all methods lose their predictive power as noted by their near-zero $R_2$ scores.

**Estimated Probability Distribution:**
- The estimated PDFs for parameter h are presented in Figure 2 and Figure 3. In lines with the above metrics, for PDFs estimated by GPR are closest to the true PDF (as noted by low Bhattacharya Distance in the metrics Tables) in the less variable training regime (Figure 2); and performance deteriorates in the high variability regime (Figure 3). The stacked-LSTM performs the worst in all experiments barring one (X, Y=8, F=20, MTU>15) which is hard to qualify as statistically significant.

**Error Growth in Predicted X Trajectory**
- (Averaged over different sampled h values) For 5<MTU<7 training regime (Figure 4), all models show low error growth in $X_1$ for MTU upto 1. In addition, for GP, the error growth amplitude remains the lowest through the MTU range considered.
- In the MTU > 15 training regime, error growth is lowest for the less chaotic case (Figure 5 (A) and (C), F=10). It reinforces the above results that GPs exhibits superior predictive power in the less temporal variable and relatively low training chaotic regimes. However, for high chaoticity, error growth for GP converges to FC-DNN and stacked-LSTM.

**Estimated Parameter h with Uncertainty Bounds**
- To compare the performance of GP alone across different experiment cases, the estimated parameter h for 80 test data points is plotted in Figure 6. As observed by the increasing width of uncertainty bounds, the predictive power of GPs decline as chaoticity increases (Figure 6[A] and 6[B]) and worsens further when complemented with a more variable training regime (Figure 6[C]).

Table 2: X=4, Y=4, F=10, Training Regime: 5 < MTU < 7

| Model | MSE | MAE | $R^2$ | Bhatta. Distance | Corr. |
|---|---|---|---|---|---|
| GP | 0.124 | 0.265 | 0.642 | 0.565 | 0.807 |
| FC-DNN | 0.247 | 0.410 | 0.291 | 1.124 | 0.549 |
| Stacked-LSTM | 0.329 | 0.482 | 0.058 | 0.835 | 0.284 |
| LR | 0.242 | 0.389 | 0.313 | 1.165 | 0.573 |

Table 3: X=4, Y=4, F=20, Training Regime: 5 < MTU < 7

| Model | MSE | MAE | $R^2$ | Bhatta. Distance | Corr. |
|---|---|---|---|---|---|
| GP | 0.119 | 0.249 | 0.656 | 0.524 | 0.819 |
| FC-DNN | 0.247 | 0.404 | 0.290 | 1.183 | 0.541 |
| Stacked-LSTM | 0.333 | 0.487 | 0.044 | 1.216 | 0.254 |
| LR | 0.252 | 0.406 | 0.530 | 1.222 | 0.288 |

Yadav et al.

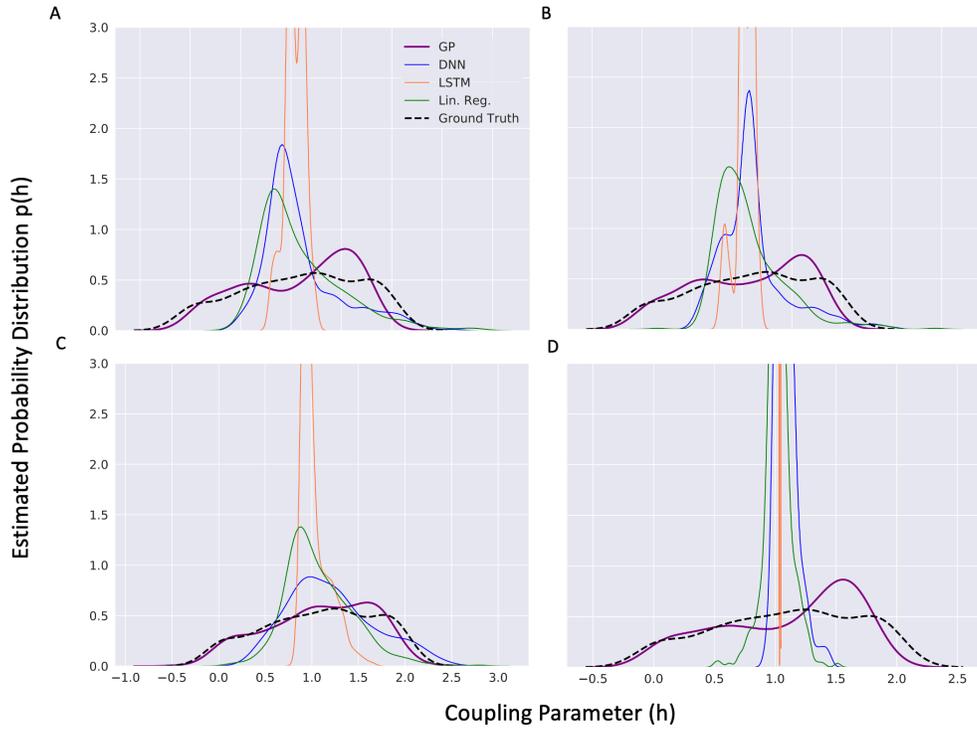

**Figure 2: Estimated Probability Distribution of coupling parameter 'h' [6] (see Eq. 1) for the 4 Experiments in Training Regime 5<MTU<7. [A] X=4, Y=4, F=10 [B] X=4, Y=4, F=20 [C] X=8, Y=8, F=10 [D] X=8, Y=8, F=20. (1 MTU = 1/Δ$t$ timesteps)**

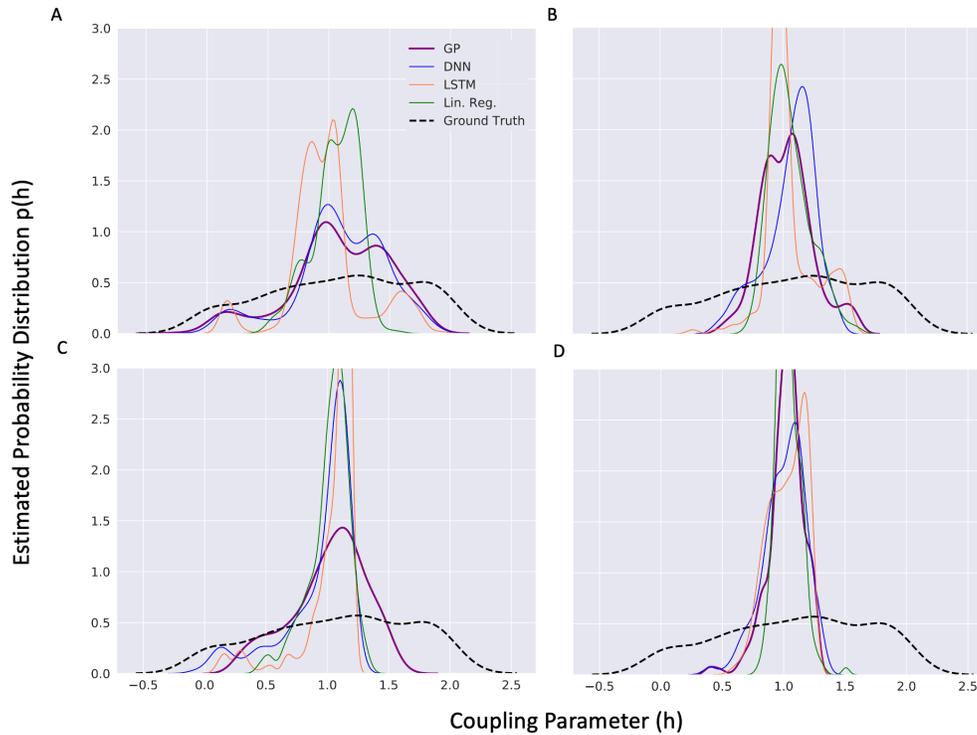

**Figure 3: Estimated Probability Distribution of coupling parameter 'h' for the 4 Experiments in Training Regime 5<MTU<7. [A] X=4, Y=4, F=10 [B] X=4, Y=4, F=20 [C] X=8, Y=8, F=10 [D] X=8, Y=8, F=20. (1 MTU = 1/Δ$t$ timesteps)**

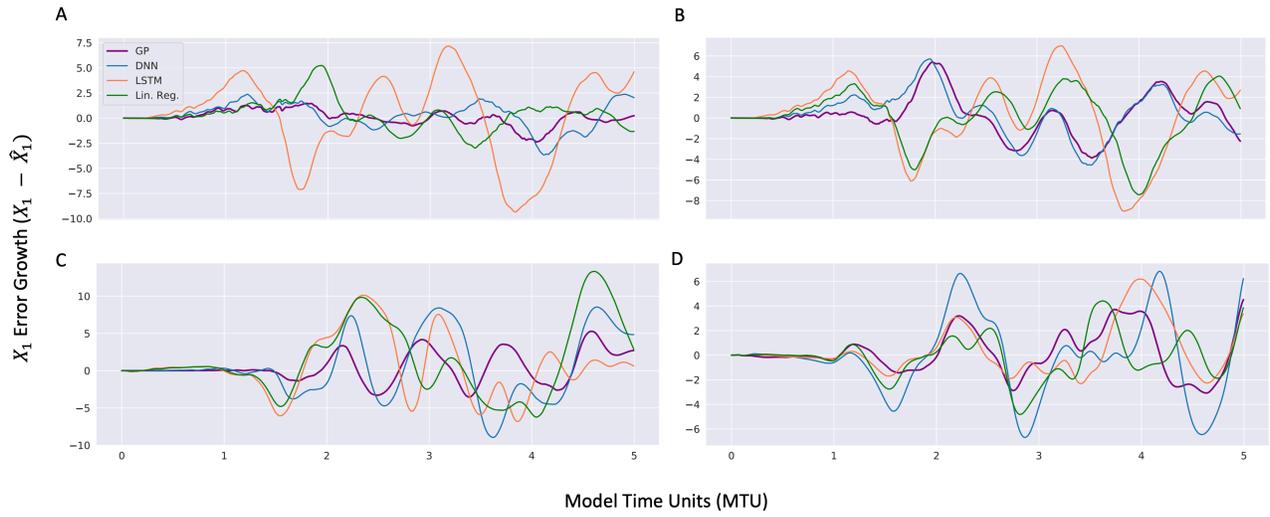

**Figure 4: Error Growth in Predicted Trajectory of $X_1$ for the 4 Experiments in Training Regime 5 < MTU < 7.**
**[A] X=4, Y=4, F=10; [B] X=4, Y=4, F=20; [C] X=8, Y=8, F=10 [D] X=8, Y=8, F=20. (1 MTU = 1/$\Delta t$ timesteps)**

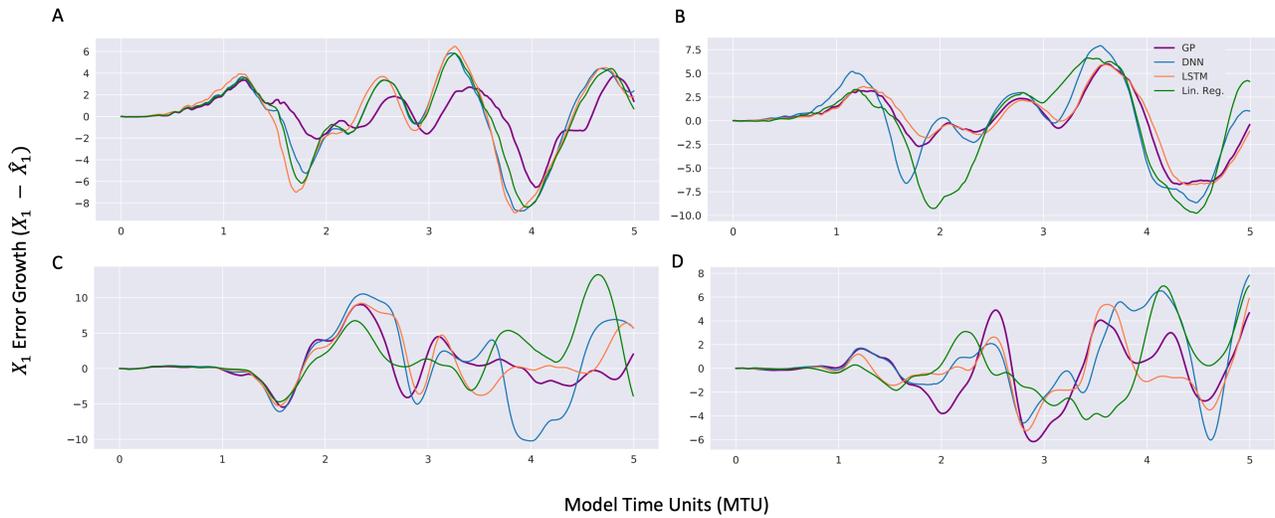

**Figure 5: Error Growth in Predicted Trajectory of $X_1$ for the 4 Experiments in Training Regime MTU < 7.**
**[A] X=4, Y=4, F=10; [B] X=4, Y=4, F=20; [C] X=8, Y=8, F=10 [D] X=8, Y=8, F=20. (1 MTU = 1/$\Delta t$ timesteps)**

**Table 4: X=8, Y=8, F=10, Training Regime: 5 < MTU < 7**

| Model | MSE | MAE | $R^2$ | Bhatta. Distance | Corr. |
|---|---|---|---|---|---|
| **GP** | **0.185** | **0.292** | **0.553** | **0.585** | **0.871** |
| **FC-DNN** | 0.205 | 0.374 | 0.411 | 0.725 | 0.681 |
| **Stacked-LSTM** | 0.348 | 0.501 | 0.154 | 1.265 | 0.453 |
| **LR** | 0.255 | 0.398 | 0.266 | 1.100 | 0.524 |

**Table 5: X=8, Y=8, F=20, Training Regime: 5 < MTU < 7**

| Model | MSE | MAE | $R^2$ | Bhatta. Distance | Corr. |
|---|---|---|---|---|---|
| **GP** | **0.164** | **0.298** | **0.516** | **0.669** | **0.804** |
| **FC-DNN** | 0.315 | 0.472 | 0.096 | 1.070 | 0.389 |
| **Stacked-LSTM** | 0.351 | 0.502 | 0.005 | 1.053 | 0.049 |
| **LR** | 0.359 | 0.506 | 0.001 | 1.080 | 0.048 |

Yadav et al.**Table 6: X=4, Y=4, F=10, Training Regime: MTU>15**

| Model | MSE | MAE | R² | Bhatta. Distance | Corr. |
|---|---|---|---|---|---|
| **GP** | **0.247** | **0.393** | **0.290** | 0.755 | **0.556** |
| **FC-DNN** | **0.253** | **0.403** | **0.274** | 0.878 | **0.534** |
| **Stacked-LSTM** | 0.308 | 0.442 | 0.117 | 0.110 | 0.410 |
| **LR** | 0.321 | 0.479 | 0.079 | 0.833 | 0.288 |

**Table 7: X=4, Y=4, F=20, Training Regime: MTU>15**

| Model | MSE | MAE | R² | Bhatta. Distance | Corr. |
|---|---|---|---|---|---|
| **GP** | **0.296** | **0.452** | 0.148 | 0.845 | 0.398 |
| **FC-DNN** | **0.299** | **0.462** | 0.117 | 0.813 | 0.363 |
| **Stacked-LSTM** | 0.313 | 0.465 | **0.100** | 1.120 | 0.324 |
| **LR** | 0.332 | 0.480 | 0.048 | 0.944 | 0.230 |

**Table 8: X=8, Y=8, F=10, Training Regime: MTU>15**

| Model | MSE | MAE | R² | Bhatta. Distance | Corr. |
|---|---|---|---|---|---|
| **GP** | **0.270** | **0.419** | **0.225** | 0.670 | **0.490** |
| **FC-DNN** | **0.279** | **0.426** | **0.199** | 0.909 | **0.518** |
| **Stacked-LSTM** | 0.297 | 0.445 | 0.147 | 1.155 | 0.392 |
| **LR** | 0.360 | 0.501 | -0.033 | 0.830 | 0.102 |

**Table 9: X=8, Y=8, F=20, Training Regime: MTU>15**

| Model | MSE | MAE | R² | Bhatta. Distance | Corr. |
|---|---|---|---|---|---|
| **GP** | 0.348 | 0.499 | **0.020** | 0.821 | **0.191** |
| **FC-DNN** | 0.352 | **0.489** | -0.009 | 0.794 | 0.160 |
| **Stacked-LSTM** | 0.353 | 0.492 | -0.007 | **0.610** | 0.134 |
| **LR** | **0.346** | 0.500 | 0.006 | 1.179 | 0.128 |

## 7 Discussion and Future Work

Parametrization schemes in climate models are one of the biggest sources of uncertainty in climate projections and accurate estimation of these parameters can translate to more accurate future predictions. Taking forward the general problem defined in [6], we apply Gaussian Processes for parameter estimation on the canonical Lorenz-96 NLD system representative of the atmospheric behavior. The intrinsic Bayesian treatment in GPs [52] also provides tools for uncertainty quantification. Where deep learning methods become ill-suited under the small data constraint, GP Regression offers a viable data-efficient learning approach as noted in the various performance metrics across different experiments.

Future work entails scaling GPs for larger datasets [43] and then making a comparison with state-of-the-art deep learning methods. Recent works in Deep GPs [44] show exciting results combining the expressiveness of deep neural nets and the ability of GPs to encode prior (physically guided) information through appropriate design of kernels.

It is pertinent to point out the "no free lunch theorem" [20] which argues that no one model performs best for all possible situations. Nearly all machine learning algorithms make certain assumptions (learning bias) about the predictor and the target value. Further, infinite width single-layer neural networks are known to be equivalent to a GP [26]. In our case, the superior performance of GP can be largely attributed to the fact the target variable is indeed jointly Gaussian and this information is hardcoded into the model through an appropriate kernel function. We observe that as training data becomes "noisier," GPR loses its predive power considerably as the underlying assumption may not hold as tightly [34]. On the other hand, deep learning models which may search over a larger solution space do not show such deterioration [45]. Although we have used the same kernel for both training regimes considered, a natural way forward is adaptive learning by using different kernel functions or even completely different learning machines for different training data.

## ACKNOWLEDGEMENT

Funding for this work was provided by Northeastern University, the US National Science Foundation's (NSF) BIGDATA Award (1447587), INQUIRE Award (1735505) and CyberSEES Award (1442728). Support from ONR grant N00014-19-1-2273 is gratefully acknowledged. The first author would like to thank fellow PhD student Kate Duffy of Northeastern University for helpful discussions.## REFERENCES

[1] Krizhevsky, A., Sutskever, I. & Hinton, G. E. ImageNet Classification with Deep Convolutional Neural Networks. in *Advances in Neural Information Processing Systems 25* (eds. Pereira, F., Burges, C. J. C., Bottou, L. & Weinberger, K. Q.) 1097–1105 (Curran Associates, Inc., 2012).
[2] Amodei, D. *et al.* Deep Speech 2: End-to-End Speech Recognition in English and Mandarin. in *International Conference on Machine Learning* 173–182 (2016).
[3] He, K., Zhang, X., Ren, S. & Sun, J. Deep Residual Learning for Image Recognition. in *2016 IEEE Conference on Computer Vision and Pattern Recognition (CVPR)* 770–778 (IEEE, 2016). doi:10.1109/CVPR.2016.90.
[4] Brunton, S. L., Proctor, J. L. & Kutz, J. N. Discovering governing equations from data by sparse identification of nonlinear dynamical systems. *PNAS* **113**, 3932–3937 (2016).
[5] G. Valenza et al., "Predicting Mood Changes in Bipolar Disorder Through Heartbeat Nonlinear Dynamics," in IEEE Journal of Biomedical and Health Informatics, vol. 20, no. 4, pp. 1034-1043, July 2016.
[6]. Schneider, T., Lan, S., Stuart, A. & Teixeira, J. Earth System Modeling 2.0: A Blueprint for Models That Learn From Observations and Targeted High-Resolution Simulations. *Geophysical Research Letters* **44**, 12,396-12,417 (2017).
[7] Lorenz, Edward N. "Designing Chaotic Models." *Journal of the Atmospheric Sciences* 62, no. 5 (2005): 1574–87. https://doi.org/10.1175/jas3430.1.
[8] Stephen Rasp, S. Online learning as a way to tackle instabilities and biases in neural network parameterizations. *arXiv:1907.01351 [physics]* (2019).

[9] O'Gorman, P. A. & Dwyer, J. G. Using machine learning to parameterize moist convection: potential for modeling of climate, climate change and extreme events. *J. Adv. Model. Earth Syst.* **10**, 2548–2563 (2018).

[10] Bryson, A., Ho, Y.C.: Applied Optimal Control. Hemisphere Publishing Corporation (1975)

[11] Gelb, Arthur. *Applied Optimal Estimation*. Cambridge, Massachusetts: The M.I.T. Press, 1994.

[12] Pathak, J., Hunt, B., Girvan, M., Lu, Z. & Ott, E. Model-Free Prediction of Large Spatiotemporally Chaotic Systems from Data: A Reservoir Computing Approach. *Phys. Rev. Lett.* **120**, 024102 (2018).

[13] Vlachas, P. R. *et al.* Forecasting of Spatio-temporal Chaotic Dynamics with Recurrent Neural Networks: a comparative study of Reservoir Computing and Backpropagation Algorithms. *arXiv:1910.05266 [physics]* (2019).

[14] Vlachas, P. R., Byeon, W., Wan, Z. Y., Sapsis, T. P. & Koumoutsakos, P. Data-Driven Forecasting of High-Dimensional Chaotic Systems with Long Short-Term Memory Networks. *Proc. R. Soc. A* **474**, 20170844 (2018).

[15] Teng, Q. & Zhang, L. Data driven nonlinear dynamical systems identification using multi-step CLDNN. *AIP Advances* **9**, 085311 (2019).

[16] J. Schoukens and L. Ljung, "Nonlinear System Identification: A User-Oriented Road Map," in *IEEE Control Systems Magazine*, vol. 39, no. 6, pp. 28-99, (2019). doi: 10.1109/MCS.2019.2938121.

[17] Raissi, M. & Karniadakis, G. E. Hidden physics models: Machine learning of nonlinear partial differential equations. *Journal of Computational Physics* **357**, 125–141 (2018).

[18] Press, T. M. Gaussian Processes for Machine Learning | The MIT Press

[19] Gentine, P., Pritchard, M., Rasp, S., Reinaudi, G. & Yacalis, G. Could Machine Learning Break the Convection Parameterization Deadlock? *Geophysical Research Letters* **45**, 5742–5751 (2018).

[20] D. H. Wolpert and W. G. Macready, "No free lunch theorems for optimization," in IEEE Transactions on Evolutionary Computation, vol. 1, no. 1, pp. 67-82, April 1997.

[21] VoosenJul. 26, P., 2018 & Pm, 2:00. Science insurgents plot a climate model driven by artificial intelligence. *Science | AAAS* (2018).

[22] Reichstein, M. *et al.* Deep learning and process understanding for data-driven Earth system science. *Nature* **566**, 195 (2019).

[23] IPCC, 2014: Climate Change 2014: Synthesis Report. Contribution of Working Groups I, II and III to the Fifth Assessment Report of the Intergovernmental Panel on Climate Change [Core Writing Team, R.K. Pachauri and L.A. Meyer (eds.)]. IPCC, Geneva, Switzerland, 151 pp.

[24] Zelinka, M. D. *et al.* Causes of Higher Climate Sensitivity in CMIP6 Models. *Geophysical Research Letters* **47**, e2019GL085782 (2020)

[25] Climate Models Are Running Red Hot, and Scientists Don't Know Why. Bloomberg.com (2020).

[26] Neal, Radford M. Bayesian learning for neural networks. Vol. 118. Springer Science & Business Media, 2012.

[27] Evensen, G. The Ensemble Kalman Filter: theoretical formulation and practical implementation. *Ocean Dynamics* **53**, 343–367 (2003). https://doi.org/10.1007/s10236-003-0036-9

[28] Ravela, S., McLaughlin, D., Fast ensemble smoothing. Ocean Dynamics 57(2), 123–134, (2007).

[29] Arulampalam, M.S., S. Maskell, N. Gordon, and T. Clapp. "A Tutorial on Particle Filters for Online Nonlinear/Non-Gaussian Bayesian Tracking." *IEEE Transactions on Signal Processing* 50, no. 2 (2002): 174–88. https://doi.org/10.1109/78.978374.

[30] Koller, Daphne, and Nir Friedman. *Probabilistic Graphical Models: Principles and Techniques*. Cambridge, MA, 2009.

[31] Seybold, Hansjörg, Sai Ravela, and Piyush Tagade. "Ensemble Learning in Non-Gaussian Data Assimilation." *Dynamic Data-Driven Environmental Systems Science Lecture Notes in Computer Science*, 2015, 227–38. https://doi.org/10.1007/978-3-319-25138-7_21.

[32] Tagade, Piyush, and Sai Ravela. "On a Quadratic Information Measure for Data Assimilation." *2014 American Control Conference*, 2014. https://doi.org/10.1109/acc.2014.6859127.

[33] A Grossman, Long Range Temperature Forecasting Using Machine Learning, Master of Engineering Thesis, Electrical Engineering and Computer Science, 56 pp., 2020

[34] C. E. Rasmussen & C. K. I. Williams, Gaussian Processes for Machine Learning, the MIT Press, 2006, ISBN 026218253X. c 2006 Massachusetts Institute of Technology. www.GaussianProcess.org/gpml

[35] Hastie, Trevor, Jerome Friedman, and Robert Tisbshirani. The Elements of Statistical Learning: Data Mining, Inference, and Prediction. New York: Springer, 2017.

[36] Chattopadhyay, Ashesh, Pedram Hassanzadeh, and Devika Subramanian. "Data-Driven Predictions of a Multiscale Lorenz 96 Chaotic System Using Machine-Learning Methods: Reservoir Computing, Artificial Neural Network, and Long Short-Term Memory Network." *Nonlinear Processes in Geophysics* 27, no. 3 (2020): 373–89. https://doi.org/10.5194/npg-27-373-2020.

[37] Arnold, H. M., I. M. Moroz, and T. N. Palmer. "Stochastic Parametrizations and Model Uncertainty in the Lorenz '96 System." *Philosophical Transactions of the Royal Society A: Mathematical, Physical and Engineering Sciences* 371, no. 1991 (2013): 20110479. https://doi.org/10.1098/rsta.2011.0479.

[38] Yuval, Janni, and Paul A. O'Gorman. "Stable Machine-Learning Parameterization of Subgrid Processes for Climate Modeling at a Range of Resolutions." *Nature Communications* 11, no. 1 (2020). https://doi.org/10.1038/s41467-020-17142-3.

[39] Mcfarlane, Norman. "Parameterizations: Representing Key Processes in Climate Models without Resolving Them." *Wiley Interdisciplinary Reviews: Climate Change* 2, no. 4 (2011): 482–97. https://doi.org/10.1002/wcc.122.

[40] Hourdin, Frédéric, Thorsten Mauritsen, Andrew Gettelman, Jean-Christophe Golaz, Venkatramani Balaji, Qingyun Duan, Doris Folini, et al. "The Art and Science of Climate Model Tuning." *Bulletin of the American Meteorological Society* 98, no. 3 (2017): 589–602. https://doi.org/10.1175/bams-d-15-00135.1.

[41] Meehl, Gerald A., Catherine A. Senior, Veronika Eyring, Gregory Flato, Jean-Francois Lamarque, Ronald J. Stouffer, Karl E. Taylor, and Manuel Schlund. "Context for Interpreting Equilibrium Climate Sensitivity and Transient Climate Response from the CMIP6 Earth System Models." *Science Advances* 6, no. 26 (2020). https://doi.org/10.1126/sciadv.aba1981.

[42] Rasp, Stephen. Lorenz '96 is too easy! Machine learning research needs a more realistic toy model. Retrieved from https://raspstephan.github.io/blog/lorenz-96-is-too-easy/

[43] Hensman, James, Nicolo Fusi, and Neil D. Lawrence. "Gaussian processes for big data." *arXiv preprint arXiv:1309.6835* (2013).

[44] Damianou, Andreas, and Neil Lawrence. "Deep gaussian processes." *Artificial Intelligence and Statistics*. 2013.

[45] Sejnowski, Terrence J. "The Unreasonable Effectiveness of Deep Learning in Artificial Intelligence." *Proceedings of the National Academy of Sciences*, 2020, 201907373. https://doi.org/10.1073/pnas.1907373117.

[46] Foreman-Mackey, Daniel, et al. "Fast and scalable Gaussian process modeling with applications to astronomical time series." *The Astronomical Journal* 154.6 (2017): 220.

[47] Blumer, Anselm, et al. "Occam's razor." *Information processing letters* 24.6 (1987): 377-380.

[48] Tangirala, Arun K. "Principles of System Identification," (2018). https://doi.org/10.1201/9781315222509.

[49] Tartakovsky, A. M., C. Ortiz Marrero, Paris Perdikaris, G. D. Tartakovsky, and D. Barajas-Solano. "Physics-Informed Deep Neural Networks for Learning Parameters and Constitutive Relationships in Subsurface Flow Problems." *Water Resources Research* 56, no. 5 (2020). https://doi.org/10.1029/2019wr026731.

[50] Qian, Yun, Charles Jackson, Filippo Giorgi, Ben Booth, Qingyun Duan, Chris Forest, Dave Higdon, Z. Jason Hou, and Gabriel Huerta. "Uncertainty Quantification in Climate Modeling and Projection." *Bulletin of the American Meteorological Society* 97, no. 5 (2016): 821–24. https://doi.org/10.1175/bams-d-15-00297.1.

[51] Bauer, Matthias, Mark van der Wilk, and Carl Edward Rasmussen. "Understanding probabilistic sparse Gaussian process approximations." *Advances in neural information processing systems*. 2016.

[52] Bilionis I., Zabaras N. Bayesian Uncertainty Propagation Using Gaussian Processes. In: Ghanem R., Higdon D., Owhadi H. (eds) Handbook of Uncertainty Quantification. Springer, Cham. (2015) https://doi.org/10.1007/978-3-319-11259-6_16-1

[53] Schneider, Tapio, João Teixeira, Christopher S. Bretherton, Florent Brient, Kyle G. Pressel, Christoph Schär, and A. Pier Siebesma. "Climate Goals and Computing the Future of Clouds." *Nature Climate Change* 7, no. 1 (2017): 3–5. https://doi.org/10.1038/nclimate3190.

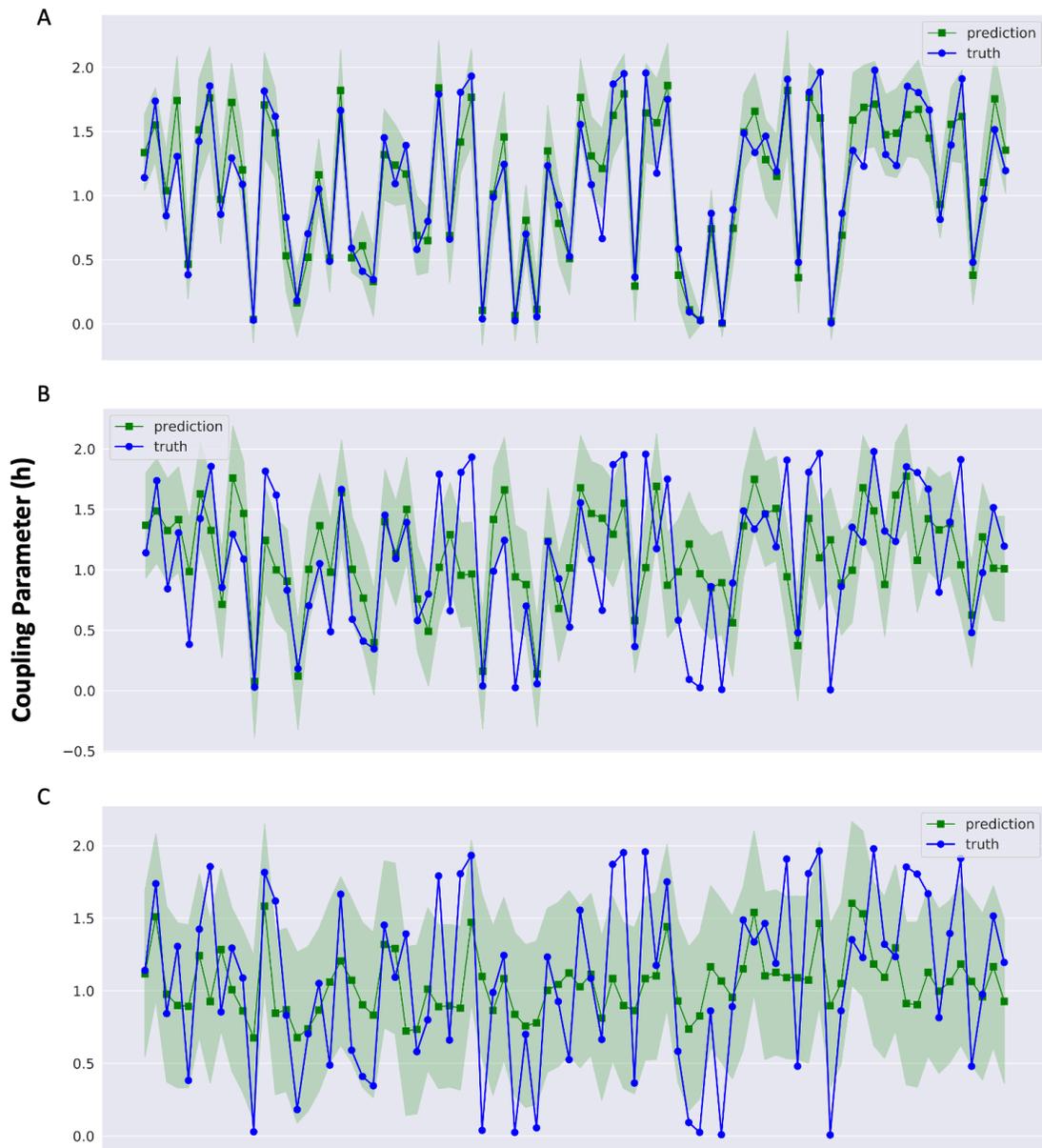

**Figure 6: Estimated coupling parameter 'h' [6] (with 95% confidence bounds) by GP Regression. X=4, Y=4. [A] F=10, Training Regime: 5<MTU<7 [B] same as [A] except F=20 [C] F=20, Training Regime: MTU>15. (1 MTU = $1/\Delta t$ timesteps)**